\pdfoutput=1                                              
\documentclass[letterpaper, 10 pt, conference]{ieeeconf}  

\IEEEoverridecommandlockouts                              

\usepackage{graphicx}
\graphicspath{{figs/}}
\usepackage{amsmath}
\usepackage{amssymb}
\usepackage{algorithm}
\usepackage{algpseudocode}
\usepackage{times}
\usepackage{booktabs}
\usepackage{multirow}
\usepackage{hhline}
\usepackage{pifont}
\usepackage{makecell}
\usepackage{dblfloatfix}
\usepackage{hyperref}

\newcommand{\xmark}{\ding{55}}

\newcommand{\TODO}[1]{}

\title{\LARGE \bf
VLAff: Vision-Language-Affordance Model \\ for Unified Actionable Affordances
}

\author{Jihoon Oh$^{1}$, Kento Kawaharazuka$^{1}$, and Kei Okada$^{1}$%
\thanks{\raggedright $^{1}$The authors are with the University of Tokyo, Japan.
        {\tt\small \{oh, kawaharazuka, k-okada\}@jsk.imi.i.u-tokyo.ac.jp}}%
\thanks{\raggedright\tiny Accepted for publication in the 2026 IEEE/RSJ International Conference on
        Intelligent Robots and Systems (IROS). \copyright{} 2026 IEEE. Personal use of this
        material is permitted. Permission from IEEE must be obtained for all other uses, in any
        current or future media, including reprinting/republishing this material for advertising
        or promotional purposes, creating new collective works, for resale or redistribution to
        servers or lists, or reuse of any copyrighted component of this work in other works.}%
}

\begin{document}

\maketitle
\thispagestyle{empty}
\pagestyle{empty}

\begin{abstract}
\indent Learning manipulation skills from human videos is promising for scalable robot learning.
  However, the embodiment mismatch between humans and robots makes this challenging.
  One promising solution is to learn object-centric actionable affordances that are embodiment-agnostic.
  In this work, we propose a framework that leverages egocentric human videos with state-of-the-art 3D Structure-from-Motion and hand mesh reconstruction to extract actionable affordances such as visual, grasp, and trajectory affordances that explicitly encode where to interact, how to grasp, and how to move.
  We construct \textbf{EgoAffordance}, a large-scale dataset comprising 204K episodes with 5.6M visual affordances and 11.6M grasp and trajectory affordances.
  Building on this, we introduce \textbf{VLAff}, a large vision-language model-based unified foundation model that learns cross-modal correlations across all actionable affordances.
  Given a visual observation and instruction, \textbf{VLAff} generates visual affordance heatmaps, grasp poses, and trajectories, which are then converted into directly executable actions by utilizing 3D scene information.
  Through extensive experiments, we demonstrate that \textbf{VLAff} not only achieves state-of-the-art performance on visual affordance prediction, but can also be effectively applied to real robot applications such as zero-shot manipulation and affordance-guided robot learning.

\end{abstract}

\begin{figure*}[t]
\centering
\includegraphics[width=\textwidth]{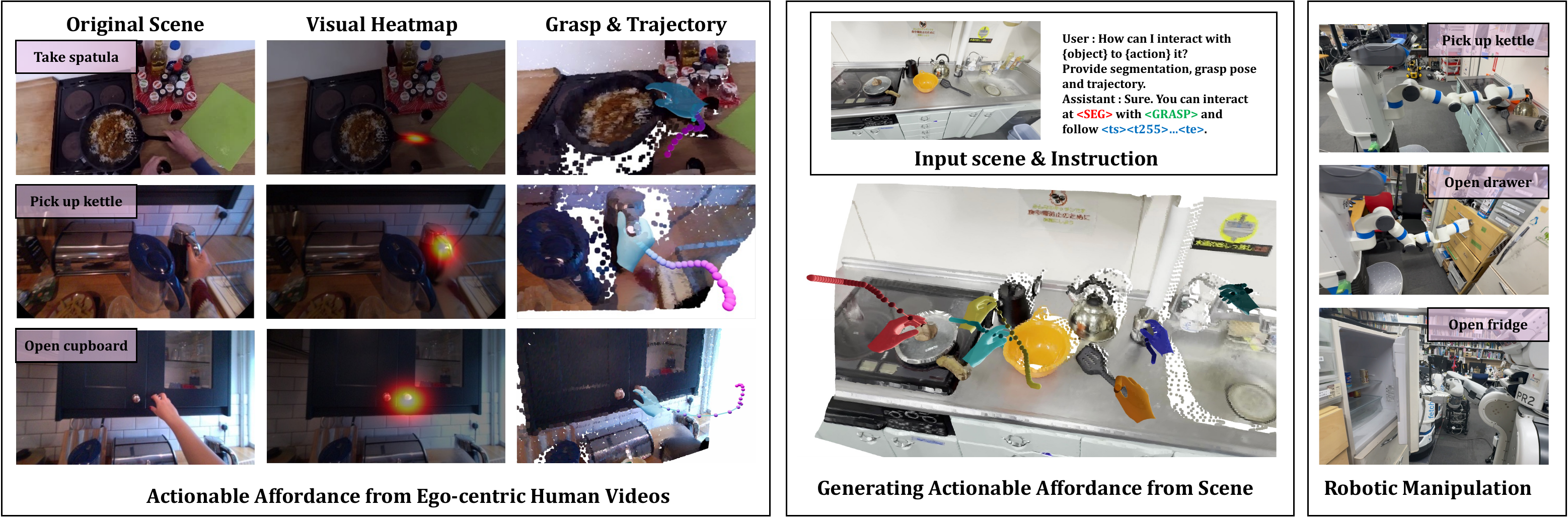}
\caption{We present \textbf{VLAff}, a Vision-Language-Affordance model that learns actionable affordances such as visual, grasp, trajectory from large-scale egocentric human videos to enable robot manipulation across diverse tasks.}
\label{fig:overview}
\end{figure*}

\section{Introduction}

\begin{table*}[!t]
\centering
\caption{Comparison of affordance-related datasets. Our EgoAffordance dataset uniquely provides comprehensive actionable affordances including visual affordance masks, hand pose annotations, and camera trajectories, extracted from large-scale egocentric human videos. Frame indicates the total number of frames, Inst (Act) denotes the number of instances or actions, and Obj represents the number of object categories.}
\label{tab:dataset_comparison}
\tiny
\resizebox{\textwidth}{!}{
\begin{tabular}{l|l|c|c|c|ccc}
\toprule
\multirow{2}{*}{\textbf{Type}} & \multirow{2}{*}{\textbf{Dataset}} & \multirow{2}{*}{\textbf{Frame}} & \multirow{2}{*}{\textbf{Inst (Act)}} & \multirow{2}{*}{\textbf{Obj}} & \multicolumn{3}{c}{\textbf{Annotation}} \\
\hhline{~|~|~|~|~|---}
\multicolumn{1}{l|}{} & \multicolumn{1}{l|}{} & \multicolumn{1}{c|}{} & \multicolumn{1}{c|}{} & \multicolumn{1}{c|}{} & Contact & Hand Pose & Camera Traj \\
\hline\hline
\multirow{6}{*}{\shortstack{Visual\\Affordance}}
& UMD~\cite{myers2015affordance} & 30K & 7 & 17 & \checkmark & \xmark & \xmark \\
& AGD20K~\cite{luo2022learning} & 23.8K & 36 & 50 & \checkmark & \xmark & \xmark \\
& IIT-AFF~\cite{nguyen2017iitaff} & 8.8K & 9 & 10 & \checkmark & \xmark & \xmark \\
& ADE-Aff~\cite{chuang2018adeaff} & 10K & 7 & 150 & \checkmark & \xmark & \xmark \\
& HOVA-500K~\cite{ma2025gloverplus} & 500K & 675 & 1.7K & \checkmark & \xmark & \xmark \\
\hline\hline
\multirow{6}{*}{\shortstack{Egocentric\\HOI}}
& H2O~\cite{kwon2021h2o} & 571K & 36 & 8 & \xmark & \checkmark & \checkmark \\
& HOI4D~\cite{liu2022hoi4d} & 2.4M & 800 & 16 & \xmark & \checkmark & \checkmark \\
& EPIC-KITCHENS~\cite{damen2018scaling} & 11.5M & 125 & 331 & \xmark & \xmark & \checkmark \\
& HD-EPIC~\cite{doughty2025hdepic} & 4.4M & 1.2K & 17K & \xmark & \xmark & \checkmark \\
& Ego4D~\cite{grauman2022ego4d} & 3.7M & 1.7K & 4.3K & \xmark & \xmark & \checkmark \\
& Ego-Exo4D~\cite{grauman2023egoexo4d} & 1.4M & 4.5K & - & \xmark & \checkmark & \checkmark \\
\hline\hline
\multirow{3}{*}{\shortstack{\textbf{Actionable}\\\textbf{Affordance}}}
& & & & & & & \\
& \textbf{EgoAffordance (Ours)} & \textbf{5.6M} & \textbf{1.7K} & \textbf{16.4K} & \checkmark & \checkmark & \checkmark \\
& & & & & & & \\
\bottomrule
\end{tabular}
}
\end{table*}

\indent Foundation models trained on large-scale datasets have proven effective for many downstream tasks in natural language processing and computer vision, such as LLaMA~\cite{touvron2023llama} and ChatGPT~\cite{openai2024gpt4technicalreport}. In robotics, there have been attempts to build such foundation models. However, traditional paradigms like imitation learning~\cite{pomerleau1991efficient} face challenges: collecting robot action-state datasets is costly and time-consuming, with task and environment domains often limited to laboratory settings. In contrast, human video datasets~\cite{grauman2022ego4d,damen2018scaling,doughty2025hdepic,grauman2023egoexo4d} can be easily collected from diverse environments and already exist at internet scale. However, direct use of human data is challenging due to embodiment differences between robots and humans, and video data lacks explicit action-state information. Previous work has attempted to learn implicit representations~\cite{ma2023vip,xiao2022masked,nair2022r3m} from human videos for robot learning, but these are generally difficult to interpret and challenging to transfer directly into robot actions.

\indent Affordances~\cite{gibson1979ecological} offer a promising bridge between human demonstrations and robot task understanding. Many previous studies have attempted human-to-robot skill transfer using affordances such as heatmaps~\cite{bahl2023vrbaffordances,ma2024glover}, grasps~\cite{chao2021dexycb}, and trajectories~\cite{shaw2023videodex} obtained from human demonstrations. However, these attempts rely on data acquired through sophisticated sensors in static laboratory environments. Recent works learn affordances from human videos, but typically focus on only one modality~\cite{ma2025gloverplus} or are limited to 2D image space~\cite{bahl2023vrbaffordances}. There is no unified model that integrates visual, grasp, and trajectory affordances into a single framework.

\indent In this work, we focus on (1) how to extract actionable affordance data from human videos, (2) how to train large-scale affordance knowledge in a single model, and (3) how to apply that affordance knowledge to downstream tasks like robot manipulation.
We propose a pipeline that applies state-of-the-art 3D structure-from-motion to egocentric human videos to extract actionable affordance data using hand-object interaction detectors, segmentation models, and 3D hand reconstruction models, with inpainting to produce agent-agnostic scenes. As shown in Table~\ref{tab:dataset_comparison}, our EgoAffordance dataset provides comprehensive actionable affordances compared to existing datasets.
We then introduce a multimodal learning framework based on large vision-language models that simultaneously predicts visual, grasp, and trajectory affordances from visual observations and instructions.
Finally, we demonstrate effectiveness on downstream tasks such as zero-shot manipulation and affordance-guided learning.

\section{Related Works}

\subsection{Learning from Human Videos}

Human videos provide a scalable source for learning manipulation behaviors. Early approaches focused on learning visual representations through self-supervised learning~\cite{nair2022r3m,xiao2022masked,ma2023vip,majumdar2023vc1,ma2023liv}. These methods learn generalizable visual features from human videos that transfer to robotic tasks across diverse embodiments and environments. While these representation learning methods improve downstream task performance, they capture implicit visual patterns without explicit action information.

More recent work has shifted toward learning explicit action representations from human videos. Methods extract contact-based affordances~\cite{bahl2023vrbaffordances}, learn contact heatmaps and hand poses~\cite{srirama2024hrp}, or predict 3D hand trajectories~\cite{han2025vidbot} for robot pre-training and zero-shot manipulation. However, these methods focus on individual affordance modalities—visual contact points, hand poses, or trajectories—in isolation. Our work differs by jointly learning all three actionable affordances in a unified framework, enabling complete manipulation understanding from visual observations to grasp configurations and motion trajectories.

\subsection{Affordance Learning}

\paragraph{Visual Affordance.}
Visual affordances indicate where interactions occur on objects. Early works detected affordances from static images~\cite{myers2015affordance,sawatzky2017weakly,li2023locate,bahl2023vrbaffordances} but were limited to simple object categories or closed-set affordances. Recent methods leverage large-scale datasets and achieve open-vocabulary affordance reasoning~\cite{ma2024glover,ma2025gloverplus,qian2024affordancellm,tang2025uad}, grounding affordances from natural language. However, these methods focus solely on visual interaction regions without modeling grasp configurations or motion trajectories.

\paragraph{Hand-Object Interaction.}
Understanding hand-object interactions is crucial for manipulation learning. Datasets capture dexterous hand poses with RGB-D data~\cite{chao2021dexycb,fan2023arctic}, detailed hand-object contact~\cite{brahmbhatt2020contactpose}, large-scale interactions with affordance annotations~\cite{yang2022oakink}, bimanual tool use~\cite{li2024taco}, and 4D temporal modeling~\cite{liu2022hoi4d}. However, these datasets and methods focus on grasp pose and contact modeling in isolation, without integrating visual affordance prediction or trajectory generation for complete manipulation understanding.

Our work differs by jointly learning visual affordances, grasp configurations, and trajectories in a unified framework, capturing their cross-modal correlations for complete actionable affordance understanding.

\subsection{Vision-Language-Action Models}

Vision-language-action (VLA) models have transformed robotic manipulation through large-scale pretraining~\cite{brohan2022rt1,brohan2023rt2,kim2024openvla,team2024octo,vuong2023openx}. These models demonstrate that transformer architectures with vision-language pretraining achieve robust manipulation across multiple robot platforms and embodiments. The development of large-scale robot datasets has been crucial for this progress. Open X-Embodiment~\cite{vuong2023openx} aggregates diverse robot manipulation data across multiple embodiments, while DROID~\cite{khazatsky2024droid} provides 76K in-the-wild manipulation trajectories across 564 scenes. These datasets enable training generalist policies that transfer across different robot platforms.

However, VLA models require robot-specific action labels and struggle with the embodiment gap from human demonstrations. Our work addresses this by learning actionable affordances as an intermediate representation, enabling effective transfer from human videos to robot control.

\section{Affordance Extraction from Human Videos}


\subsection{Preliminaries}

We formulate the task of learning actionable affordances from human videos as a multi-modal prediction problem. Given an RGB image $\mathbf{I} \in \mathbb{R}^{H \times W \times 3}$ and a natural language instruction $\mathbf{L}$, our goal is to predict three complementary object-centric affordance representations:

\begin{equation}
f_{\text{VLAff}}(\mathbf{I}, \mathbf{L}) = (\mathbf{A}_v, \mathbf{A}_g, \mathbf{A}_t)
\end{equation}

where:
\begin{itemize}
    \item $\mathbf{A}_v \in \mathbb{R}^{H \times W}$ is the visual affordance heatmap indicating the probability distribution of related regions in the scene.
    \item $\mathbf{A}_g \in \mathbb{R}^{96}$ is the grasp pose represented as MANO parameters~\cite{romero2017mano}, capturing the hand configuration during manipulation. This includes global wrist rotation and 15 local joint rotations, each represented using 6D rotation representation~\cite{zhou20196d}.
    \item $\mathbf{A}_t \in \mathbb{R}^{T \times 6}$ is the trajectory of the hand after interaction, where each waypoint includes both translation (3D) and rotation (3D), and $T$ is the number of trajectory waypoints.
\end{itemize}

To generate object-centric actionable affordances, we first identify the peak interaction point from the visual affordance heatmap. We then project this 2D point to 3D space using the camera intrinsics and depth information to obtain 3D anchor point. The grasp pose and trajectory are subsequently transformed relative to this 3D anchor point through the spatial transformation, yielding the object-centric grasp pose and trajectory that align the predicted affordances to the object's coordinate frame centered at the interaction point. This object-centric formulation enables effective transfer of learned human manipulation strategies across different embodiments, as the complete affordance representation provides both the where (visual), how (grasp), and motion (trajectory) information necessary for execution.

\subsection{Actionable Affordance Generation from Ego-Centric Video}

\paragraph{Data Preparation}
We begin by processing egocentric video datasets using their provided language descriptions and temporal annotations. For each video segment with language description $\mathbf{L}$, we sample frames $\{I_t\}$ around the annotated interaction timestamps. We query a pretrained Large VLM~\cite{openai2024gpt4technicalreport} on subsampled frames to extract key interaction information: contact keyframes $\{I_c\}$, action labels, object categories, and interacting hand information.

\paragraph{Hand-Object Interaction Extraction}
For the identified contact keyframes $\{I_c\}$, we apply a hand-object detector~\cite{shan2020h2o} to obtain bounding boxes for hands and objects. We generate object masks~\cite{ravi2024sam2} and extract 2D hand keypoints~\cite{xu2022vitpose}. Contact regions are identified by computing the intersection between fingertip keypoints and object masks, then tracked~\cite{karaev2024cotracker3}.

For 3D hand reconstruction, we extract MANO parameters~\cite{li2024wilor} representing hand pose and shape. To minimize visual bias from the ego perspective, we perform ego segmentation~\cite{zhang2022egohos} followed by inpainting~\cite{zhou2023propainter}.

\paragraph{Ego-centric SfM}
To reconstruct the 3D scene and trajectory information, we perform monocular depth estimation~\cite{wang2025moge2} on the inpainted frames. For videos lacking camera intrinsics, we estimate them~\cite{schoenberger2016sfm}. We then estimate camera poses~\cite{teed2021droid}, filtering out dynamic regions using object and ego masks to ensure robust estimation.

With the estimated camera poses and depth maps, we track 3D hand trajectories~\cite{harley2024tapip3d}, capturing both pre- and post-contact motion patterns essential for trajectory affordance learning.

\section{Vision-Language-Affordance Model}

\begin{figure*}[ht]
\centering
\includegraphics[width=0.95\textwidth]{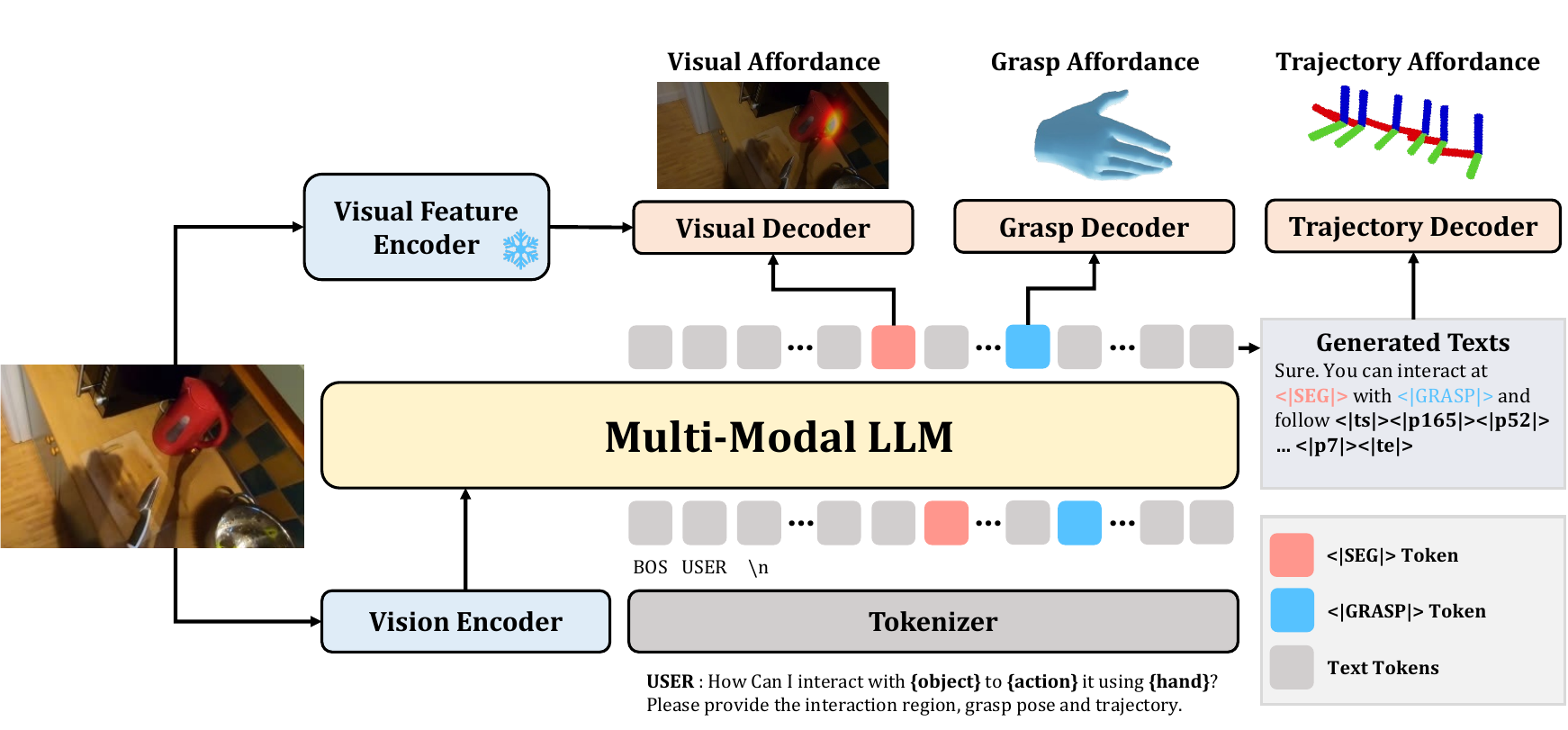}
\caption{VLAff model architecture showing the integration of vision encoder, VLM, and specialized decoders for visual, grasp, and trajectory affordance prediction.}
\label{fig:model_architecture}
\end{figure*}

To enable unified learning of visual, grasp, and trajectory affordances, we extend a pre-trained vision-language model (VLM) with specialized affordance tokens. Our approach introduces three types of tokens to represent different affordance modalities:

\textbf{Visual Affordance Token.} We add a special token \texttt{<SEG>} to the VLM vocabulary to predict visual affordance heatmaps. Following the approach in LISA~\cite{lai2024lisa}, the \texttt{<SEG>} token embedding is used to generate pixel-wise affordance probabilities through a segmentation decoder, producing the visual affordance heatmap $\mathbf{A}_v$.

\textbf{Grasp Token.} For grasp pose prediction, we introduce a \texttt{<GRASP>} token that encodes MANO hand parameters. The model learns to associate language instructions with corresponding hand configurations by fine-tuning the VLM to predict the 96-dimensional grasp affordance $\mathbf{A}_g \in \mathbb{R}^{96}$, which includes global wrist rotation and 15 local joint rotations represented using 6D rotation representation~\cite{zhou20196d}.

\textbf{Trajectory Tokens.} To handle trajectories autoregressively as discrete tokens, we employ spatial binning~\cite{brohan2022rt1,brohan2023rt2,kim2024openvla} to quantize the trajectory space. We discretize the 6D pose space (3D translation + 3D rotation) into discrete bins and create a vocabulary of trajectory tokens \texttt{<p0>}, ..., \texttt{<p$_{N-1}$>}, where each token corresponds to a specific bin in the quantized trajectory space. During training, continuous trajectory waypoints are converted to their nearest bin indices, and the model learns to predict sequences of trajectory tokens that can be decoded back to continuous 6D poses.

By introducing these specialized affordance tokens, the model enables joint learning across all three affordance modalities and captures rich correlations between visual interaction regions, grasp configurations, and motion patterns.

\subsection{Model Architecture}

Our VLAff model consists of four main components that work together to predict comprehensive affordances from vision-language inputs. Figure~\ref{fig:model_architecture} illustrates the overall architecture.

\paragraph{VLM}
The core vision-language model processes the input image and language instruction to generate contextualized embeddings for all tokens in the sequence. The special affordance tokens \texttt{<SEG>}, \texttt{<GRASP>}, and \texttt{<p*>} are embedded within this sequence, producing corresponding token embeddings that are used by the specialized decoders.

For trajectory affordance prediction, the model generates trajectory tokens autoregressively. Given the context tokens up to position $t$, the model predicts the next trajectory token. This autoregressive generation continues until the model produces an end-of-trajectory token or reaches the maximum sequence length, resulting in a trajectory sequence that can be decoded back to continuous 6D poses.

\paragraph{Vision Encoder}
While VLMs excel at general knowledge and reasoning capabilities, they often exhibit limitations in fine-grained visual grounding tasks that require precise spatial understanding. To address this limitation and enhance visual feature extraction beyond the standard VLM vision encoder, we incorporate an additional DINOv2~\cite{oquab2023dinov2} vision encoder specifically designed for dense visual understanding. The DINOv2 encoder extracts rich visual features that provide fine-grained spatial information crucial for precise affordance localization, complementing the VLM's reasoning capabilities with stronger visual grounding abilities.

\paragraph{Visual Affordance Decoder}
The visual affordance decoder fuses the enhanced visual features $\mathbf{F}_v$ from the DINOv2 encoder with the \texttt{<SEG>} token embedding $\mathbf{h}_{seg}$ to generate the visual affordance heatmap. Through a series of upsampling and fusion layers, the decoder produces the final visual affordance map $\mathbf{A}_v \in \mathbb{R}^{H \times W}$, where each pixel value represents the probability of being an interaction region.

\paragraph{Grasp Decoder}
The grasp decoder takes the \texttt{<GRASP>} token embedding $\mathbf{h}_{grasp}$ and decodes it into MANO hand parameters. Specifically, it predicts the global wrist rotation and 15 local joint rotations using 6D rotation representation~\cite{zhou20196d}, forming the complete 96-dimensional grasp affordance $\mathbf{A}_g \in \mathbb{R}^{96}$.

\subsection{Training Objectives}

We train VLAff using specialized loss functions tailored to each affordance modality:

\paragraph{Visual Affordance Loss}
For visual affordance prediction, we employ the Soft Dice Loss to handle the imbalanced nature of visual heatmap prediction tasks:
\begin{equation}
\mathcal{L}_{visual} = 1 - \frac{2 \sum_{i,j} (\mathbf{A}_v^{pred}(i,j))^p \cdot (\mathbf{A}_v^{gt}(i,j))^p + s}{\sum_{i,j} (\mathbf{A}_v^{pred}(i,j))^p + \sum_{i,j} (\mathbf{A}_v^{gt}(i,j))^p + s}
\end{equation}
where $\mathbf{A}_v^{pred}$ and $\mathbf{A}_v^{gt}$ are the predicted and ground truth visual affordance heatmaps, $s=1.0$ is a smoothing factor, and $p=1.5$ is a power parameter for soft boundary handling.

\paragraph{Grasp Affordance Loss}
For grasp pose prediction, we use Smooth L1 loss:
\begin{equation}
\mathcal{L}_{grasp} = \frac{1}{J} \sum_{k=1}^{J} f(x_k), \quad x_k = \mathbf{A}_g^{pred}[k] - \mathbf{A}_g^{gt}[k]
\end{equation}
where $J=96$ and $f(x) = 0.5x^2$ if $|x| < 1$, else $|x| - 0.5$.

\paragraph{Trajectory Affordance Loss}
For trajectory token prediction, we apply the standard Cross-Entropy loss used in autoregressive language models:
\begin{equation}
\mathcal{L}_{traj} = -\frac{1}{T} \sum_{t=1}^{T} \log P(\texttt{<pi>}_t^{gt} | \mathbf{I}, \mathbf{L}, \texttt{<p*>}_{1:t-1})
\end{equation}
where $\texttt{<pi>}_t^{gt}$ is the ground truth trajectory token at position $t$.

\paragraph{Total Loss}
The total training loss combines all three objectives:
\begin{equation}
\mathcal{L}_{total} = \lambda_v \mathcal{L}_{visual} + \lambda_g \mathcal{L}_{grasp} + \lambda_t \mathcal{L}_{traj}
\end{equation}
where $\lambda_v$, $\lambda_g$, and $\lambda_t$ are weighting hyperparameters that balance the contribution of each affordance modality during training.

\subsection{Trajectory Sampling Strategy}

Our model infers trajectories from 2D images and language instructions, which poses challenges for generating physically plausible 3D trajectories. To address this limitation, we leverage the reasoning capabilities of our vision-language model backbone and introduce two strategies that guide the model to generate more plausible trajectories in 3D space during inference.

\paragraph{In-Context Trajectory Guidance}
We utilize high-performance VLMs such as GPT~\cite{openai2024gpt4technicalreport} as a high-level planner to provide directional guidance for trajectory generation. Given the task prompt and interaction object, we query the VLM to provide a 3D directional vector $[d_x, d_y, d_z]$ in the ego-centric (camera) coordinate frame, where each component is constrained to $\{-1, 0, 1\}$ and encodes coarse motion intent along a camera axis: $d_x$ horizontal (right $+1$), $d_y$ vertical (down $+1$), and $d_z$ depth (forward, away from camera, $+1$), with $0$ meaning no motion on that axis. For example, $[0, 0, 1]$ indicates pushing forward, while $[1, 0, -1]$ indicates pulling backward and to the right. This guidance is applied only to the first trajectory token by sampling the initial xyz position tokens around the normalized and scaled directional vector:
\begin{equation}
\mathbf{p}_1 \sim \mathcal{N}(\mathbf{p}_{contact} + s \cdot \frac{\mathbf{d}_{guide}}{\|\mathbf{d}_{guide}\|}, \sigma^2 \mathbf{I})
\end{equation}
where $\mathbf{p}_{contact}$ is the interaction point, $\mathbf{d}_{guide}$ is the VLM-provided direction vector, $s$ is a scale factor, and $\sigma$ controls the sampling variance. This approach enables our model to produce trajectories that align with high-level task understanding while maintaining fine-grained spatial details learned from human demonstrations.

\paragraph{Sampling-Based Trajectory Selection}
To generate physically plausible trajectories, we sample $K$ trajectory candidates using nucleus sampling with top-p filtering, then select the optimal one using a composite objective that balances collision avoidance and directional alignment. For collision detection, we voxelize the 3D space and check trajectory waypoints against occupied voxels from the scene point cloud and depth map. The trajectory with the lowest combined objective is selected as the final output.

\section{Experiments}

We evaluate our VLAff model across four key aspects: (1) visual affordance prediction quality to assess how well the model localizes interaction regions, (2) trajectory generation effectiveness to validate our autoregressive generation strategy, (3) zero-shot manipulation performance to demonstrate direct transferability of generated affordances to robotic systems, and (4) affordance-guided policy learning to show how predicted affordances can serve as effective guidance for training environment-adaptive closed-loop policies.

\begin{figure*}[t]
\centering
\includegraphics[width=\textwidth]{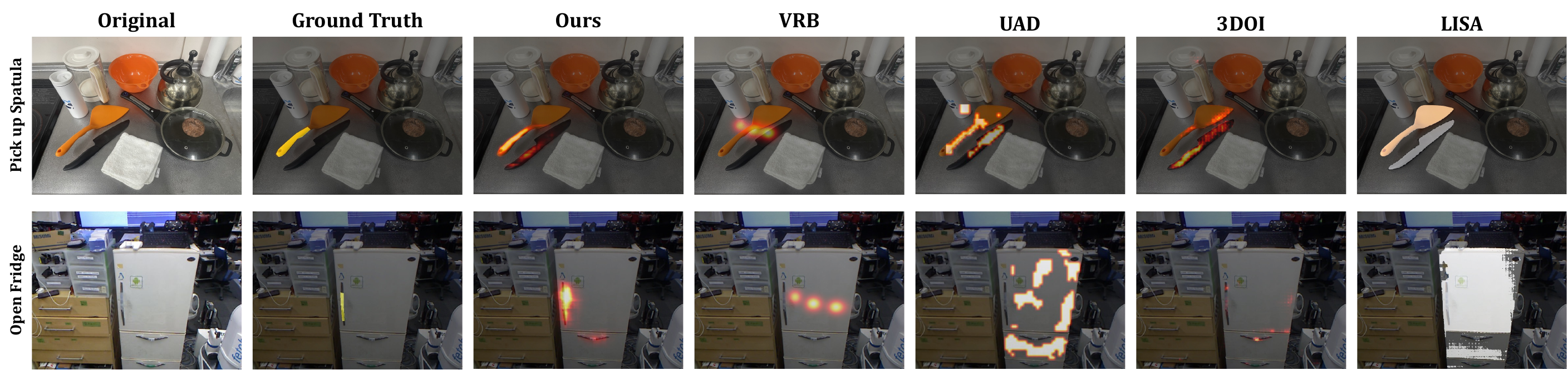}
\caption{Visual affordance prediction results in the wild. We show visual affordance predictions from VLAff and baseline methods on diverse manipulation scenarios. VLAff demonstrates superior localization of interaction regions compared to existing methods.}
\label{fig:visual_affordance_examples}
\end{figure*}

\subsection{Experimental Settings}

\paragraph{Backbones}
Our VLAff model leverages state-of-the-art foundation models as core components. We employ Qwen2.5-VL~\cite{bai2025qwen25vl} as our base vision-language model. For enhanced visual feature extraction, we integrate DINOv2~\cite{oquab2023dinov2} as our additional vision encoder.

\paragraph{Datasets}
Our final EgoAffordance dataset comprises 204,025 episodes, containing 5,782,431 visual heatmaps and 11,612,524 trajectory sequences. To improve fine-grained visual affordance prediction, we additionally incorporate data from HANDAL~\cite{guo2023handal} and SceneFun3D~\cite{delitzas2024scenefun3d}, which provide detailed object-level affordance annotations.

\paragraph{Visual Affordance}
We evaluate visual affordance prediction performance on a test set consisting of 500 randomly sampled scenes from each egocentric video dataset. We compare VLAff with state-of-the-art segmentation-based methods: VRB~\cite{bahl2023vrbaffordances}, UAD~\cite{tang2025uad}, 3DOI~\cite{qian2023doi}, and LISA~\cite{lai2024lisa}. Note that LISA does not directly predict affordance heatmaps but is an LLM-based open vocabulary segmentation model; we normalize its output segmentation masks to [0,1] for fair comparison. We employ the following evaluation metrics: \textbf{IoU (Intersection over Union)} measures the overlap between predicted and ground truth affordance regions after normalizing heatmaps to [0,1] and applying a threshold of 0.5 to create binary masks; \textbf{NSS (Normalized Scanpath Saliency)} measures how well predicted affordances align with ground truth interaction points; \textbf{SIM (Similarity)} measures the similarity between predicted and ground truth affordance distributions; \textbf{KLD (Kullback-Leibler Divergence)} quantifies the divergence between predicted and ground truth affordance probability distributions.

\paragraph{Zero-Shot Manipulation}
We evaluate VLAff's zero-shot manipulation capabilities against baseline methods on diverse manipulation tasks in both simulation and real-world environments. We compare against RAM~\cite{kuang2024ram}, VRB~\cite{bahl2023vrbaffordances}, GeneralFlow~\cite{yuan2024generalflow}, and VidBot~\cite{han2025vidbot}. For RAM, we sample retrieval episodes from our dataset. For VRB, we lift predicted 2D trajectories to 3D leveraging the strategy from RAM.

Since baseline methods do not directly output grasp poses, we employ GraspNet~\cite{fang2020graspnet} to generate robot gripper pose for all baseline methods. For VLAff, we follow the retargeting strategy from previous work~\cite{papagiannis2025rx} to estimate gripper pose from predicted hand pose.

For simulation experiments, we use IsaacGym~\cite{makoviychuk2021isaac} as our simulation platform, with tasks from PartManip~\cite{geng2023partmanip}, FrankaKitchen~\cite{gupta2019relay}, and ManiSkill~\cite{gu2023maniskill} that are included in the Ag2Manip~\cite{geng2024ag2manip} environment. We select 10 everyday household tasks that require understanding of object interaction spots and manipulation trajectories. These tasks encompass various action primitives such as opening, closing, pushing, and turning. Each method is tested 10 times per task without any task-specific training.

For real-world experiments, we deploy the generated affordances on mobile manipulation robots Fetch and PR2 to validate practical applicability. We evaluate on 5 manipulation tasks in real kitchen environments, testing each method 10 times per task.

\paragraph{Affordance-Guided Manipulation Learning}
While VLAff demonstrates strong zero-shot manipulation capabilities, there exists a gap between affordances learned from videos and physical environments in real-world deployment. To bridge this gap, we employ real-world reinforcement learning guided by the predicted affordances, where learned actionable affordances provide structured priors to accelerate policy learning in the target robot embodiment. The reward function consists of two components: (1) a sparse task success reward determined by a vision-language model (VLM) that determines task success or failure, and (2) affordance-based guidance rewards that provide dense signals throughout the manipulation based on predicted contact points and trajectories (details in supplementary material). To evaluate the effectiveness of different affordance components as guidance for manipulation learning, we conduct an ablation study on the "Open Fridge" task in a real-world environment, comparing configurations using all affordance modalities versus ablated versions removing one modality at a time.

\subsection{Experiment Results}

\subsubsection{Visual Affordance}

\begin{table}[t]
\centering
\footnotesize
\caption{Visual affordance evaluation results}
\label{tab:visual_affordance}
\begin{tabular*}{\columnwidth}{@{\extracolsep{\fill}}lcccc@{}}
\toprule
Method & IoU$\uparrow$ & NSS$\uparrow$ & SIM$\uparrow$ & KLD$\downarrow$ \\
\midrule
VRB~\cite{bahl2023vrbaffordances} & 0.013 & -0.032 & 0.052 & 6.942 \\
UAD~\cite{tang2025uad} & 0.063 & 1.095 & 0.097 & 5.687 \\
3DOI~\cite{qian2023doi} & 0.012 & 0.742 & 0.034 & 4.252 \\
LISA~\cite{lai2024lisa} & 0.113 & 1.342 & 0.025 & 3.886 \\
\midrule
\textbf{Ours} & \textbf{0.121} & \textbf{1.542} & \textbf{0.142} & \textbf{2.517} \\
\bottomrule
\end{tabular*}
\end{table}

Table~\ref{tab:visual_affordance} presents comprehensive visual affordance prediction results. VLAff achieves state-of-the-art performance across all segmentation-based metrics, demonstrating the effectiveness of our unified affordance learning approach. Beyond quantitative improvements, VLAff produces more fine-grained and natural affordance predictions compared to baseline methods, as illustrated in Figure~\ref{fig:visual_affordance_examples}. We analyze the limitations of baseline methods: VRB predicts discrete contact points rather than dense heatmaps, limiting fine-grained affordance localization; UAD, trained on rendered object images, struggles in cluttered real-world scenes; LISA, trained on object and instance-level segmentation data, tends to predict entire objects rather than specific contact points. We attribute VLAff's superior performance to two key factors: (1) the integration of DINOv2's part-aware visual features, which enable fine-grained localization of interaction regions, and (2) training on our large-scale, diverse EgoAffordance dataset spanning multiple domains and object categories. The results validate our key hypothesis that jointly learning visual, grasp, and trajectory affordances in a unified framework leads to better visual affordance prediction compared to methods that learn visual affordances in isolation.

\subsubsection{Zero-Shot Manipulation}

\begin{table*}[t]
\centering
\footnotesize
\caption{Zero-shot manipulation performance. Success rates averaged over 10 trials per task. \textbf{Sim (T01-T10):} \textbf{T01}: Open Hinge Cabinet, \textbf{T02}: Open Slide Cabinet, \textbf{T03}: Open Microwave, \textbf{T04}: Close Microwave, \textbf{T05}: Pick Up Kettle, \textbf{T06}: Open Dish Washer, \textbf{T07}: Lift Lid, \textbf{T08}: Pull Drawer, \textbf{T09}: Close Hinge Cabinet, \textbf{T10}: Turn Faucet. \textbf{Real (T11-T15):} \textbf{T11}: Open Drawer, \textbf{T12}: Pick Up Bucket, \textbf{T13}: Open Pot Lid, \textbf{T14}: Take Pan, \textbf{T15}: Pick Up Kettle.}
\label{tab:zero_shot}
\begin{tabular}{l|cccccccccc|c|ccccc|c}
\toprule
Method & \textbf{T01} & \textbf{T02} & \textbf{T03} & \textbf{T04} & \textbf{T05} & \textbf{T06} & \textbf{T07} & \textbf{T08} & \textbf{T09} & \textbf{T10} & \textbf{Avg} & \textbf{T11} & \textbf{T12} & \textbf{T13} & \textbf{T14} & \textbf{T15} & \textbf{Avg} \\
\midrule
RAM~\cite{kuang2024ram} & 40 & 20 & 50 & 80 & 90 & 40 & 40 & 60 & 90 & 0 & 51.0 & 60 & 20 & 20 & 20 & 0 & 24.0 \\
VRB~\cite{bahl2023vrbaffordances} & 20 & 10 & 60 & 70 & 90 & 30 & 50 & 50 & 80 & 20 & 48.0 & 60 & 40 & 20 & 20 & 0 & 28.0 \\
GFlow~\cite{yuan2024generalflow} & 10 & 10 & 60 & \textbf{100} & 90 & 60 & 70 & 50 & 90 & 10 & 55.0 & 80 & 20 & 20 & 20 & 0 & 28.0 \\
VidBot~\cite{han2025vidbot} & \textbf{70} & \textbf{80} & \textbf{80} & \textbf{100} & \textbf{100} & 70 & \textbf{100} & \textbf{100} & \textbf{100} & 50 & \textbf{85.0} & \textbf{100} & \textbf{60} & \textbf{60} & 40 & 0 & 52.0 \\
Ours & \textbf{70} & 60 & 70 & \textbf{100} & \textbf{100} & \textbf{80} & 90 & \textbf{100} & \textbf{100} & \textbf{60} & 83.0 & \textbf{100} & \textbf{60} & \textbf{80} & \textbf{60} & \textbf{40} & \textbf{68.0} \\
\bottomrule
\end{tabular}
\end{table*}

Table~\ref{tab:zero_shot} presents the zero-shot manipulation results across diverse tasks in both simulation and real-world environments. In simulation tasks, VLAff achieves an average success rate of 83.0\%, demonstrating strong generalization performance close to VidBot~\cite{han2025vidbot} (85.0\%), the current state-of-the-art zero-shot manipulation framework. VidBot's strong overall performance in simulation can be attributed to its ability to access spatial input such as depth, enabling the model to directly leverage spatial characteristics for manipulation planning, which is particularly beneficial in controlled simulation environments.

Notably, VLAff demonstrates superior performance on real-world tasks, achieving an average success rate of 68.0\%, 16 percentage points higher than VidBot's 52.0\%. This performance gap validates our key design choice of training on large-scale real-world egocentric video datasets, which better capture the complexity and diversity of real manipulation scenarios, while other methods frequently fail to estimate appropriate interaction regions in cluttered real-world scenes. VLAff's unified affordance learning approach enables more robust contact interaction prediction through joint learning of visual, grasp, and trajectory affordances. Our failure cases primarily stem from incorrect initial direction sampling during trajectory generation, which can be improved through better integration of scene context and spatial reasoning.

\begin{figure}[t]
\centering
\includegraphics[width=\columnwidth]{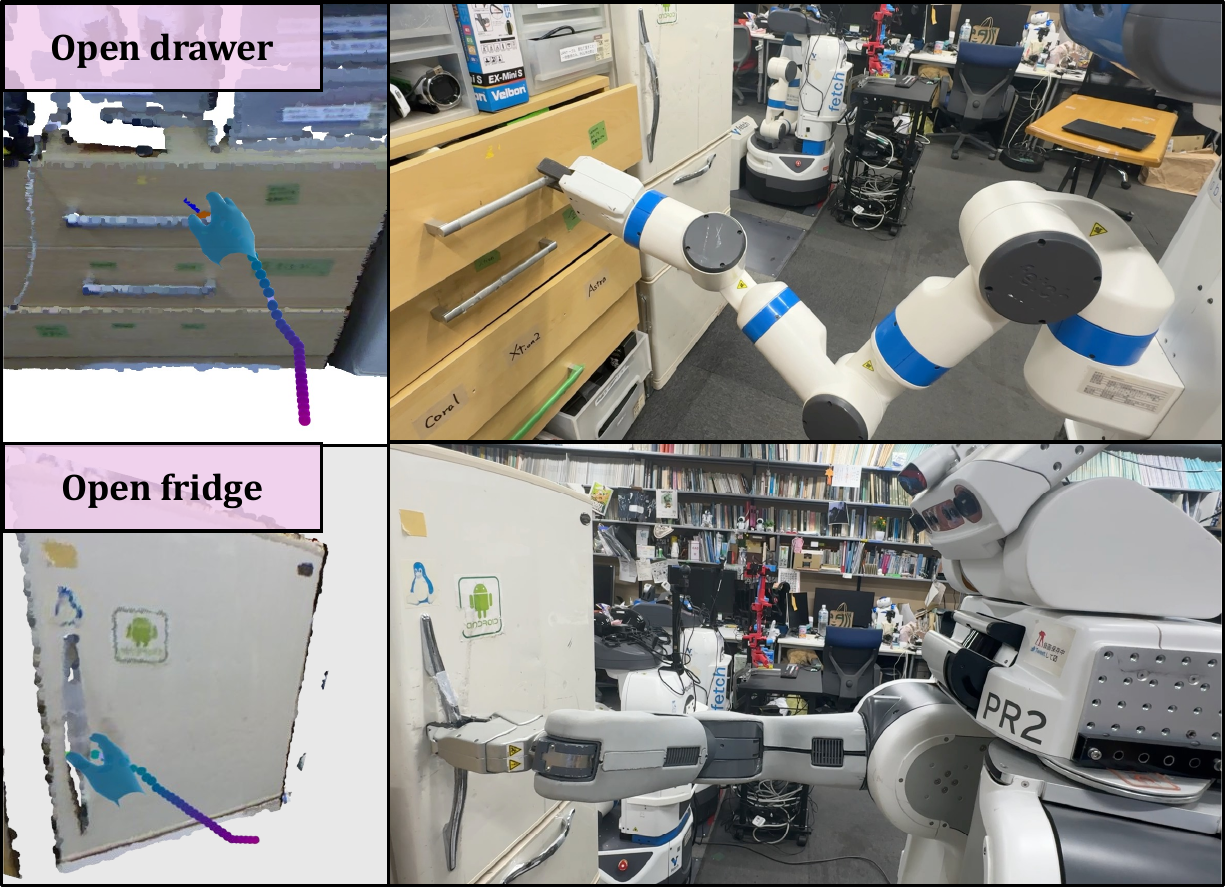}
\caption{Robot manipulation experiments. We visualize the generated affordances and their application to real robot manipulation tasks.}
\label{fig:robot_manipulation}
\end{figure}

\subsubsection{Affordance-Guided Manipulation Learning}

\begin{figure}[t]
\centering
\includegraphics[width=0.48\textwidth]{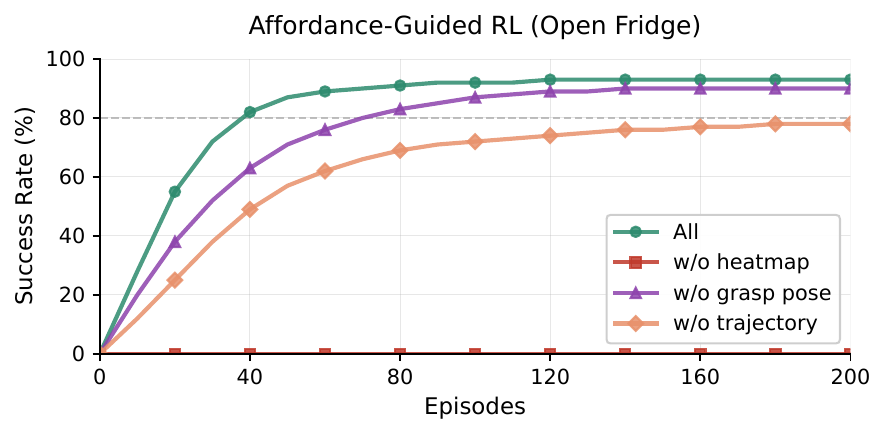}
\caption{Affordance-guided manipulation learning results on real-world Open Fridge task. We compare learning efficiency when using different combinations of affordance guidance}
\label{fig:affordance_rl}
\end{figure}

Figure~\ref{fig:affordance_rl} shows the learning curves for each configuration. The complete model with all affordance modalities achieves the best performance most efficiently, validating that visual heatmap, grasp pose, and trajectory information provide complementary guidance. Most critically, when the visual affordance heatmap is removed, the model completely fails to achieve successful manipulation regardless of the number of training episodes, demonstrating that visual affordance is absolutely essential for manipulation learning as it provides the most important cues for object interaction by indicating where to make contact. Interestingly, grasp pose plays a less significant role in manipulation learning, which we attribute to the fundamental morphological differences between human hands and robot grippers—the exact hand configuration from human demonstrations does not directly transfer to gripper-based manipulation.

\section{Conclusion}

We built EgoAffordance, a large-scale actionable affordance dataset comprising visual heatmaps, grasp poses and manipulation trajectories extracted from egocentric human videos leveraging state-of-the-art computer vision techniques.
Upon this dataset, we presented VLAff, a novel Vision-Language-Affordance model that, unlike previous methods treating visual, grasp, and trajectory affordances separately, provides comprehensive action guidance by jointly learning these modalities within a unified VLM framework to predict actionable affordances from language instructions.
We demonstrate through extensive experiments on visual affordance prediction, zero-shot manipulation, and affordance-guided manipulation learning that VLAff achieves results comparable to or surpassing existing state-of-the-art methods, validating the effectiveness of our unified affordance learning approach for robotic manipulation tasks.
While VLAff demonstrates strong performance, a limitation remains: occasional generation of implausible trajectories that may not respect physical constraints.
Future work could address this through incorporation of 3D scene-aware model architectures~\cite{xu2024pointllm}, as demonstrated in recent work~\cite{yoshida2025egoscaler}.

\bibliographystyle{IEEEtran}
\bibliography{main}

\end{document}